\documentclass{article}
\usepackage{spconf,amsmath,amssymb,graphicx,setspace}
\usepackage{arydshln}
\usepackage{amssymb}
\usepackage{bm}
\usepackage{algorithm}
\usepackage{algorithmic}
\usepackage{textcomp}
\usepackage{tipa}
\usepackage{cite} 
\usepackage{url}
\usepackage{float}
\usepackage{tikz}      
\usepackage{xcolor}    
\usepackage{amsthm}
\usepackage{amsmath}

\usepackage[caption=false,font=footnotesize]{subfig}
\usepackage{booktabs}
\usepackage[table]{xcolor}
\usepackage{makecell}
\definecolor{lightyellow}{RGB}{255,245,184}
\definecolor{lightgraycol}{RGB}{238,238,238}
\title{ADAPTIVE SHARED EXPERTS WITH LORA-BASED MIXTURE OF EXPERTS FOR MULTI-TASK LEARNING}
\name{Minghao Yang \quad Ren Togo \quad Guang Li \quad Takahiro Ogawa \quad Miki Haseyama \thanks{
This research was supported in part by JP23K11141, JP23K21676, JP24K02942, JP24K23849, and JP25K21218.}
\address{Hokkaido University \\
    \{yang, togo, guang, ogawa, mhaseyama\}@lmd.ist.hokudai.ac.jp}}
\begin{document}
\ninept
\maketitle

\begin{abstract}
Mixture-of-Experts (MoE) has emerged as a powerful framework for multi-task learning (MTL). However, existing MoE–MTL methods often rely on single-task pretrained backbones and suffer from redundant adaptation and inefficient knowledge sharing during the transition from single-task to multi-task learning (STL-to-MTL). To address these limitations, we propose adaptive shared experts (ASE) within a low-rank adaptation (LoRA) based MoE, where shared experts are assigned router-computed gating weights jointly normalized with sparse experts. This design facilitates STL-to-MTL transition, enhances expert specialization, and cooperation. Furthermore, we incorporate fine-grained experts by increasing the number of LoRA experts while proportionally reducing their rank, enabling more effective knowledge sharing under a comparable parameter budget. Extensive experiments on the PASCAL-Context benchmark, under unified training settings, demonstrate that ASE consistently improves performance across diverse configurations and validates the effectiveness of fine-grained designs for MTL.
\end{abstract}
\par
\section{Introduction}
Large-scale deep neural networks have driven significant advances in various domains~\cite{jiao2024multiscale, 10937907}, including speech~\cite{speech1, wang-etal-2024-massive, meng2025dolphin}, vision~\cite{ChenAdaMVMoEAdaptiveMultiTask2023, jin2024moh,TeamGraniteVisionLightweight2025}, and natural language processing~\cite{DeepSpeedMoE2022, DaiDeepSeekMoEUltimateExpert2024,DouLoRAMoEAlleviateWorld2024}. A key technique of these advances is the Mixture-of-Experts (MoE) architecture~\cite{lepikhin2021gshard, DaiDeepSeekMoEUltimateExpert2024,jin2024moh}, which expands model capacity while keeping manageable computation by activating only a small subset of experts for each input.  Motivated by this, recent studies have explored MoE for multi-task learning (MTL)~\cite{LiangM3ViTMixtureofExpertsVision2022,ChenAdaMVMoEAdaptiveMultiTask2023,ChenModSquadDesigningMixtures2023,YangMultiTaskDensePrediction2024}, where different tasks are trained jointly to exploit both shared representations and task-specific specialization.

Despite these advantages, existing MoE--MTL approaches typically built upon a single-task learning (STL) pretrained backbone. To serve multiple tasks, this backbone requires shifting from task-specific to task-agnostic representations (STL-to-MTL). Due to sparse and varying activation, each expert is compelled to learn this transition independently. This results in inefficient adaptation, where different experts may redundantly relearn overlapping knowledge, while task-agnostic features are insufficiently captured. Consequently, both efficiency and effective task specialization are compromised.

A natural solution is to incorporate a shared expert~\cite{DeepSpeedMoE2022, DaiDeepSeekMoEUltimateExpert2024} dedicated to capturing common knowledge for the STL-to-MTL transition, thereby relieving task-specific experts from redundant adaptation and enabling finer specialization. However, shared experts in MoE-MTL face critical limitations. First, common implementations add the shared expert’s output directly to that of sparsely activated experts, implicitly assigning it a fixed, dominant gating weight~\cite{DeepSpeedMoE2022, DaiDeepSeekMoEUltimateExpert2024}. This introduces an imbalance between the shared and sparse experts. 
Second, because the shared expert is jointly optimized across all tasks, its disproportionately large influence often amplifies gradient conflicts in later training stages~\cite{VandenhendeMultiTaskLearningDense2021}. Moreover, MoE–MTL models are also expensive to train, since activating and routing multiple experts increases the memory footprint and parameter updates even under sparse computation~\cite{DouLoRAMoEAlleviateWorld2024,YangMultiTaskDensePrediction2024}.

To overcome these challenges, we propose a new MoE-MTL framework with three innovations. First, we introduce an adaptive shared expert (ASE) whose contribution is governed by a router-computed gating weight and normalized together with the selected sparse experts. This design achieves stable and balanced outputs through joint normalization. During the transition of STL-to-MTL, the shared experts initially take the lead but gradually reduce their influence, which helps alleviate gradient conflicts. Second, to ensure computational efficiency, we implement each expert as a low-rank adaptation module (LoRA)~\cite{hu2022lora}, substantially reducing parameter and FLOP overhead while preserving expressivity. Third, we introduce the notion of fine-grained experts~\cite{fine_grained_moe,DaiDeepSeekMoEUltimateExpert2024}: by increasing the number of LoRA experts while proportionally lowering their rank, we maintain a comparable parameter budget but achieve finer specialization and more effective cooperation across tasks.
We evaluate our method on the PASCAL-Context~\cite{pascal} MTL benchmark. Experimental results demonstrate the robustness and effectiveness of our framework across different settings, as well as the benefits of fine-grained configurations for improving multi-task performance.

Our contributions are summarized as follows.
\begin{itemize}
    \item We introduce the first ASE design for MTL, enabling more effective STL-to-MTL transfer, reducing gradient conflicts, and enhancing expert specialization.
    \item We propose an efficient LoRA–MoE design for MTL that preserves expressivity while substantially reducing computational and parameter overhead.
    \item We demonstrate that fine-grained experts provide better specialization and cooperation in MTL, leading to improved performance and stability.
\end{itemize}

\section{Methodology}
\subsection{Preliminaries}
\textbf{Mixture-of-Experts.} 
MoE~\cite{DaiDeepSeekMoEUltimateExpert2024,DouLoRAMoEAlleviateWorld2024,DeepSpeedMoE2022} duplicates a sub-module, such as the feed-forward network (FFN) in a Transformer block, into a set of $N$ parallel experts and activates only a small subset per input. Formally, for each input $\bm{\bm{x}}\in\mathbb{R}^{d_{\text{in}}}$, a router computes gating logits over all experts and derives gating scores as:
\begin{equation}
    \bm{g}=\text{Softmax}(\bm{\mathrm{W}}_g \bm{x}) \in \mathbb{R}^N,\quad \bm{\mathrm{W}}_g \in \mathbb{R}^{N\times d_{\text{in}}},
    \label{gate_score}
\end{equation}
where $\bm{g}$ denotes the gating scores and $\bm{\mathrm{W}}_g$ are trainable router parameters. 
The router then selects the top-$k$ experts and computes the final output as:
\begin{equation}
    \bm{y}=\bm{x}+\sum_{i\in\mathcal{T}} \bm{g}_iE_i(\bm{x}),\quad \mathcal{T} = \mathrm{TopK}(\bm{g}, k),
\end{equation}
where $E_i$ is the $i$-th expert in the selected set $\mathcal{T}$ and $\bm{g}_i$ is its gating score.
Although only a few experts are activated per input, all experts must be stored and updated during training, resulting in considerable memory and computation overhead~\cite{DouLoRAMoEAlleviateWorld2024,YangMultiTaskDensePrediction2024}.

\textbf{LoRA variant of MoE (LoRA-MoE).} LoRA-MoE~\cite{DouLoRAMoEAlleviateWorld2024,YangMultiTaskDensePrediction2024} is a parameter-efficient variant of MoE, where each expert is replaced by a low-rank adaptation (LoRA) module~\cite{hu2022lora} added to the frozen base weights:
\begin{equation}
    \bm{y}=\bm{x}+\bm{\mathrm{W}}_0\bm{x}+\sum_{i\in\mathcal{T}}\bm{g}_iE_i(\bm{x}),
\end{equation}
where $\bm{\mathrm{W}}_0 \in \mathbb{R}^{d_{\text{out}} \times d_{\text{in}}}$ denotes frozen base weights. Here, each expert is implemented as a LoRA module:
\begin{equation}
    E(\bm{x})= \Delta \bm{\mathrm{W}x} = \bm{\mathrm{BA}x},
\end{equation}
where the update matrix $\Delta \bm{\mathrm{W}} \in \mathbb{R}^{d_{\text{out}} \times d_{\text{in}}}$ is factorized into two low-rank matrices $\bm{\mathrm{A}}\in \mathbb{R}^{r\times d_{\text{in}}}$ and $\bm{\mathrm{B}}\in \mathbb{R}^{d_{\text{out}} \times r}$, with rank $r \ll \min(d_{\text{in}},d_{\text{out}})$. 
This design substantially reduces trainable parameters and FLOPs while preserving the flexibility of MoE, making it suitable for MTL scenarios. Hence, in this work, we build upon and extend LoRA-MoE for MTL.

\subsection{Adaptive Shared Expert} \label{sec.ase}
Existing MoE-MTL frameworks require the backbone to adapt from task-specific to task-agnostic representations. Because MoE activates experts sparsely, each expert must learn this STL-to-MTL transition in isolation, which leads to redundant adaptation and insufficient capture of common features. This inefficiency compromises both overall effectiveness and expert specialization.

\textbf{Shared Expert with LoRA-MoE.} 
The shared expert~\cite{DeepSpeedMoE2022, DaiDeepSeekMoEUltimateExpert2024} was introduced to capture and consolidate common knowledge across tasks, thus relieving task-specific experts from redundant adaptation and enabling finer specialization and better cooperation across tasks. 
When combined with LoRA-MoE, the $N$ experts include $S$ shared experts $E^s$, whose outputs are directly added as:
\begin{equation}
    \bm{y}=\bm{x}+\bm{\mathrm{W}}_0\bm{x}+\sum_{i\in\mathcal{T}}\bm{g}_iE_i(\bm{x})+\sum^{S}_{i=1}E^s_i(\bm{x}).
\end{equation}
This direct addition is equivalent to adding $\sum_{i=1}^{S} \bm{g}^s_iE^s_i(\bm{x})$ with fixed gating scores $\bm{g}^s_i=1$ for each shared expert. Since the gating scores of sparse experts sum to $\sum_{i\in\mathcal{T}}\bm{g}_i \leq 1$ after softmax (Eq.~\ref{gate_score}), the total shared contribution $\sum^{S}_{i=1} \bm{g}^s_i = S$ dominates the mixture, creating imbalance between sparse and shared experts. 
Moreover, this imbalance also disrupts the scale between the base weights $\bm{\mathrm{W}}_0$ and the LoRA experts $\Delta \bm{\mathrm{W}}$, since $\sum_{i\in\mathcal{T}} \bm{g}_i + \sum^{S}_{i=1} \bm{g}^s_i > S$. 
In MTL settings, where shared experts are trained across all tasks, such dominant and fixed contributions further aggravate gradient conflicts in later training stages.

\begin{figure}[t]
    \centering
    \includegraphics[width=0.95\linewidth]{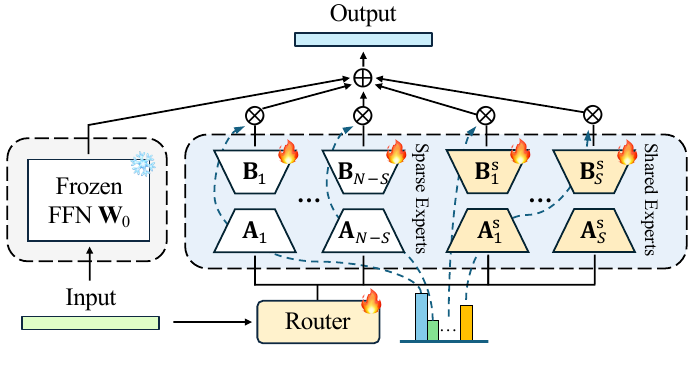}\hfill
    \includegraphics[width=0.65\linewidth]{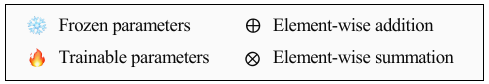}
    \vspace{-2mm}
\caption{
An illustration of the proposed adaptive shared experts. 
The frozen FFN $\bm{\mathrm{W}}_0$ is combined with sparse experts $E_i(\cdot;\,\bm{\mathrm{A}}_i,\bm{\mathrm{B}}_i)$  and adaptive shared experts $E_i^s(\cdot;\,\bm{\mathrm{A}}^{s}_i,\bm{\mathrm{B}}^{s}_i)$, whose contributions are assigned adaptive gating weights computed by task-specific routers and normalized jointly with sparse experts.
}

    \label{fig:ase}
    \vspace{-3mm}
\end{figure}

\textbf{Proposed Adaptive Shared Expert.} 
To address these problems, we propose adaptive shared experts (ASE), as illustrated in Fig.~\ref{fig:ase}, whose contribution is dynamically computed by the router and normalized together with sparse experts. 
Specifically, we compute logits for sparse and shared experts separately:
\begin{align}
    &\bm{z} = \bm{\mathrm{W}}_g \bm{x} \in \mathbb{R}^{N-S}, 
    && \bm{\mathrm{W}}_g \in \mathbb{R}^{(N-S)\times d_{\text{in}}}, \\
    &\bm{z}^{s} = \bm{\mathrm{W}}_s \bm{x} \in \mathbb{R}^{S}, 
    && \bm{\mathrm{W}}_s \in \mathbb{R}^{S\times d_{\text{in}}},
\end{align}
where $\bm{z}$ and $\bm{z}^{s}$ denote the logits for sparse and shared experts, respectively. 
To maintain the same number of active experts, we reduce the selection to $\mathcal{T} = \mathrm{TopK}(\bm{z}, k-S)$. 
Finally, sparse and shared logits are normalized jointly:
\begin{align}
    \bm{g}^{s}_i &= 
    \frac{\exp(\bm{z}^{s}_i)}{\sum_{j \in \mathcal{T}} \exp(\bm{z}_j) + \sum_{j=1}^{S} \exp(\bm{z}^{s}_j)}, 
    && i=1,\dots,S, \\
    \bm{g}_i &= 
    \frac{\exp(\bm{z}_i)}{\sum_{j \in \mathcal{T}} \exp(\bm{z}_j) + \sum_{j=1}^{S} \exp(\bm{z}^{s}_j)}, 
    && i \in \mathcal{T}.
\end{align}
This post-softmax normalization ensures that the contributions of all activated experts sum to one, thereby balancing sparse and shared experts as well as stabilizing the scale of the LoRA modules. 
At the same time, the adaptive gating enables the shared experts to contribute more strongly during the early STL-to-MTL phase, while gradually reducing their weight in later stages, thus alleviating gradient conflicts. 
The final output is defined as:
\begin{equation}
    \bm{y}=\bm{x}+\bm{\mathrm{W}}_0\bm{x}+\sum_{i\in\mathcal{T}}\bm{g}_iE_i(\bm{x})+\sum_{i=1}^{S}\bm{g}^s_iE^s_i(\bm{x}).
\end{equation}

Overall, the proposed design balances the contributions of sparse and shared experts, mitigates gradient conflicts, and enables a more efficient STL-to-MTL transfer within the LoRA-MoE framework.

\subsection{Multi-Task Learning with LoRA-MoE}
\label{mtl_framework}

We now describe the overall framework that integrates the proposed adaptive shared expert with LoRA-MoE into a unified Transformer backbone. Our design follows a standard MTL setup, in which tasks share a common backbone while retaining task-specific components. 

\textbf{Task-specific routing and heads.} For each task $t \in \{1,\dots,T\}$, the input $\bm{x}$ is first augmented with a learnable task embedding $\bm{e}_t$, as: $\bm{x}_t = \bm{x} + \bm{e}_t$. The augmented input is then processed independently by the shared backbone. 
To enable task-specific expert selection, at each layer $\ell \in \{1,\dots,L\}$ with Transformer block $B^{(\ell)}$, each task employs its own task-specific router $R^{(\ell)}_t$ for expert routing and score computing. This ensures that experts are shared at the parameter level but activated in a task-dependent manner. 
Given the hidden state $\bm{h}^{(\ell-1)}_t$, the update of the current block is:
\begin{equation}
    \bm{h}^{(\ell)}_t = B^{(\ell)}\bigl(\bm{h}^{(\ell-1)}_t;\, R^{(\ell)}_t\bigr), 
    \qquad \bm{h}^{(0)}_t=\bm{x}_t.
\end{equation}

After passing through all $L$ layers, the hidden state is fed into a task-specific head $H_t$ to produce the prediction:
\begin{equation}
    \hat{\bm{y}}_t = H_t(\bm{h}^{(L)}_t).
\end{equation}
Overall, all tasks share the same backbone and experts, but differ in their task embeddings $\bm{e}_t$, routers $\{R^{(\ell)}_t\}_{\ell=1}^{L}$, and output heads $\{H_t\}$. 
This design enables efficient parameter sharing while retaining task-specific flexibility. For clarity, the end-to-end mapping for task $t$ can be written as:
\begin{equation}
    \hat{\bm{y}}_t \;=\; H_t\Big(F\big(\bm{x}+\bm{e}_t;\, \{R^{(\ell)}_t\}_{\ell=1}^{L}\big)\Big),
\end{equation}
where $F(\cdot;\{R^{(\ell)}_t\})$ denotes the shared ViT with LoRA--MoE layers and adaptive shared experts, routed by the task-specific routers.

\textbf{Parameterization strategy.}  For efficiency, the backbone is initialized from a pretrained STL model and fine-tuned using LoRA. 
Specifically, all experts are implemented as LoRA modules, and LoRA updates are also applied to the backbone layers. 
In contrast, routers $\{R^{(\ell)}_t\}_{\ell=1}^{L}$ and task-specific heads $\{H_t\}$ are trained from scratch in full-parameter mode. 
This parameterization ensures the majority of parameters remain lightweight, while the routing and prediction components maintain full expressivity.

\section{Experiments} 
\subsection{Experiment Setup}\label{sec:exp_setting}
\textbf{Dataset and Evaluation Metrics.} We evaluate the effectiveness of our MTL framework on PASCAL-Context~\cite{pascal}, which contains 10,103 images with five task annotations of edge detection (\emph{Edge}), semantic segmentation (\emph{Seg.}), human parts segmentation (\emph{H.Parts}), surface normals (\emph{Norm.}), and saliency detection (\emph{Sal.}). 

For each task, we adopt standard evaluation metrics: mean intersection over union (mIoU)~\cite{EveringhamPascalVisualObject2010} for \emph{Seg.}, \emph{H.Parts}, and \emph{Sal.}; mean error (mErr)~\cite{mse} for \emph{Norm.}; and the optimal dataset F-measure (odsF)~\cite{5557884} for \emph{Edge}. 
Since computing odsF involves dense thresholding and is computationally expensive, we additionally report the Balanced Cross-Entropy Loss (BCE)~\cite{7410521} for \emph{Edge} in the ablation study. 
Following previous works~\cite{LiangM3ViTMixtureofExpertsVision2022, ChenModSquadDesigningMixtures2023}, we adopt the average relative performance drop $\Delta_m$ to evaluate an MTL model $m$ with respect to the baseline STL model $b$ over all tasks: $  \Delta_m = \tfrac{1}{T} \sum_{t=1}^{T} (-1)^{l_t}\,\frac{M_{m,t} - M_{b,t}}{M_{b,t}},$ where $M_{m,t}$ and $M_{b,t}$ denote the evaluation metric of task $t$ for the MTL and STL models, respectively, and $l_t=1$ if a lower value indicates better performance. 

\textbf{MoE Loss.} 
We employ the Mod-Squad loss~\cite{LiangM3ViTMixtureofExpertsVision2022} to regularize the routers. 
This loss maximizes the mutual information between tasks and experts, thereby encouraging sparse yet strong task–expert dependencies. 
Such a constraint promotes more effective expert specialization and complements our adaptive shared expert design.

\textbf{Implementation Details.} 
We integrate adaptive shared experts (ASE) with LoRA-MoE into each feed-forward network (FFN) layer of a ViT-small backbone pretrained on STL~\cite{dosovitskiy2020vit}. 
Each layer contains 16 LoRA experts with rank $r=4$, including a single adaptive shared expert, and we set the top-$k$ to 3. 
LoRA with rank 4 is applied to the backbone for parameter-efficient training, while newly added modules (routers and task-specific heads) are trained in full. 
All experiments are conducted on a single NVIDIA RTX 6000 Ada GPU. 
To reduce memory usage, all experiments are conducted with a reduced input resolution of $224\times224$ for 40 training epochs. 

\textbf{Fine-Grained Expert Settings.} 
To further enhance expert specialization and cooperation in MTL, we adopt the notion of fine-grained experts~\cite{fine_grained_moe,DeepSpeedMoE2022}, where the number of experts is increased while their LoRA rank is proportionally reduced to maintain a comparable parameter budget. 
We consider three levels of configurations, denoted as $(N/k/S/r)$, where $N$ is the total number of experts per layer, $k$ is the number of sparsely activated experts, $S$ is the number of adaptive shared experts, and $r$ is the LoRA rank of each expert. 
The $(16/3/1/4)$ configuration corresponds to the initial setup, while $(32/6/2/2)$ and $(64/12/4/1)$ represent medium- and high-granularity variants. 
Although decreasing $r$ balances the parameter count of experts, the router cost still grows with $N$, making $(64/12/4/1)$ more computationally demanding. 
Thus, we mainly report results for $(16/3/1/4)$ and $(32/6/2/2)$, and include $(64/12/4/1)$ only as a supplementary comparison.

\begin{table*}[t]
  \centering
  \caption{Comparisons with MTL methods on the PASCAL-Context dataset. 
  Best results are highlighted in \textbf{bold}.}
  \label{tab:pascal_context_results}
  \setlength{\tabcolsep}{10pt}
  \renewcommand{\arraystretch}{0.85} 
  \resizebox{0.97\textwidth}{!}{%
  \begin{tabular}{l|c|ccccc|>{\columncolor{gray!0}}c|c}
    \toprule
    Method & Backbone &
    \makecell{\emph{Seg.} \\ (mIoU)$\uparrow$} &

    \makecell{\emph{Norm.} \\ (mErr)$\downarrow$} &
    \makecell{\emph{H. Parts} \\ (mIoU)$\uparrow$} &
    \makecell{\emph{Sal.} \\ (mIoU)$\uparrow$} &
    \makecell{\emph{Edge} \\ (odsF)$\uparrow$} &
    \makecell{$\Delta_m$ \\ (\%)$\uparrow$} &
    \makecell{Params \\ (M)$\downarrow$} \\
    \midrule
    STL & ResNet-18 & 60.1 & \textbf{15.4} & 51.0 & 63.2 & 50.8 & 0.00 & \textbf{11} \\
    MTL & ResNet-18 & 58.4 & 15.9 & 51.8 & 62.6 & 50.3 & -1.29 & \textbf{11} \\
    MTAN~\cite{mtan} & ResNet-18 & 58.9 & 16.4 & 51.9 & 62.2 & 50.6 & -1.74 & \textbf{11} \\
    NDDR-CNN~\cite{nddr_cnn} & ResNet-18 & 59.7 & 15.5 & 51.7 & 62.7 & 51.5 & +0.13 & \textbf{11} \\
    Cross-Stitch~\cite{cross_sritch} & ResNet-18 & 60.4 & \textbf{15.4} & 51.6 & 63.1 & 51.7 & +0.66 & \textbf{11} \\
    \midrule
    MTL-ViT & ViT-base & 69.1 & 16.2 & 54.8 & 61.9 & 49.9 & +2.68 & 104 \\
    Baseline (16/4/0/4) & LoRA-MoE ViT-base & 73.7 & 17.5 & 59.2 & 62.9 & 53.7 & +6.06 & 107 \\
    Baseline (32/8/0/2) & LoRA-MoE ViT-base & 73.8 & 17.4 & 59.3 & 62.9 & 53.3 & +6.11 & 108 \\
    \midrule
    \textbf{Ours} (16/3/1/4) & LoRA-MoE ViT-base & \textbf{74.0} & 17.3 & 60.1 & 63.2 & \textbf{55.3} & +7.49 & 107 \\
    \textbf{Ours} (32/6/2/2) & LoRA-MoE ViT-base & \textbf{74.0} & {17.2} & \textbf{60.3} & \textbf{63.3} & 54.9 & \textbf{+7.58} & 108 \\
    \bottomrule
  \end{tabular}}
    \vspace{-1mm}
\end{table*}

\subsection{Evaluation Results}
We now present the experimental results of our framework. For each experiment, we denote the vanilla LoRA-MoE as our baseline. For ablation studies, we report BCE for \emph{Edge} and compute $\Delta_m$ against a ViT-base STL model. 
For comparisons with other MTL methods, we adopt odsF for \emph{Edge} and compute $\Delta_m$ with respect to an STL model based on ResNet-18, following common practice.

\begin{figure}[t]
  \centering
       \includegraphics[width=0.6\linewidth]{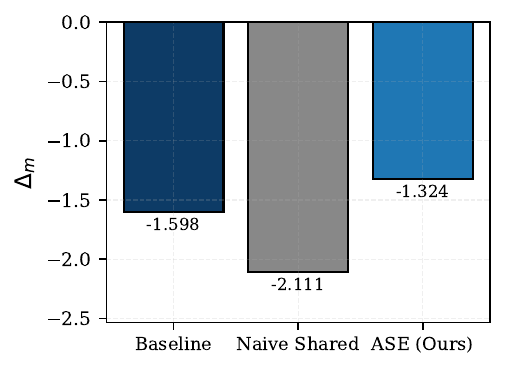}\hspace{1mm}
    \vspace{-2mm}
  \caption{\textbf{Comparison between baseline, naive shared expert, 
    and the proposed ASE.} BCE is used as the evaluation metric for \emph{Edge}. 
     The $\Delta_m$ is computed with respect to a ViT-based STL model.}\label{fig:mod_vs_ase_exp}
       \vspace{-2mm}
\end{figure}

\textbf{Comparison With the Naive Shared Expert.} 
We first compare our ASE with the baseline (vanilla LoRA-MoE), and its variant with a naive shared expert. 
As shown in Fig.~\ref{fig:mod_vs_ase_exp}, the naive shared expert leads to clear performance degradation due to imbalance and gradient conflict issues (Sec.~\ref{sec.ase}). 
In contrast, ASE reverses this trend with consistent improvement. 
These results highlight the effectiveness of ASE in not only mitigating the limitations of the naive design but also unlocking the potential of shared experts in MTL.

\textbf{Ablation on the Number of Activated Experts.} We vary the number of sparsely activated experts $k \! \in \! \{3,4,5,6,7\}$ while fixing the number of shared experts $S$ to 1. As shown in Fig. 3(a), our ASE consistently yields higher $\Delta_m$ than the baseline across all $k$, indicating robustness to the choice of top-$k$.
Moreover, when comparing under the same {total activated experts} $k_{\text{tot}}{=}k{+}S$, ASE still outperforms the baseline.

\begin{figure}[t] 
  \vspace{-1mm}
  \centering
  \subfloat[Performance across top-${k}$.]{
    \includegraphics[width=0.495\linewidth]{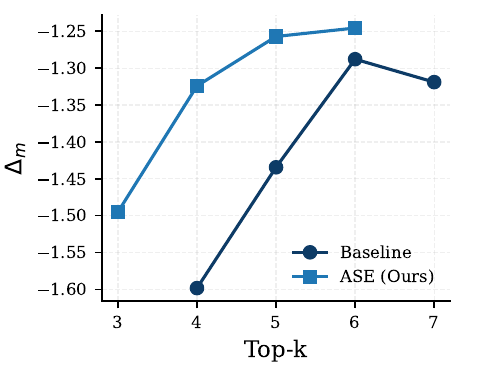}
    \label{fig:mod_vs_ase_k}
  }
  \subfloat[Fine-grained expert settings.]{
  \includegraphics[width=0.49\linewidth]{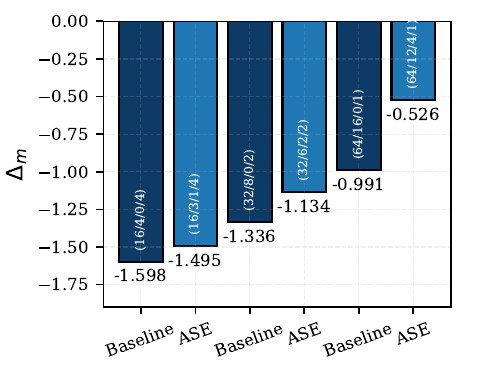}
    \label{fig:mod_vs_ase_fine}
  }
  \caption{\textbf{ASE under different top-$\bm{k}$ and fine-grained settings.} We adopt BCE for \emph{Edge}, and compute $\Delta_m$ relative to a ViT-based STL model.
  (a) Line plot of $\Delta_m$ versus top-$k$ with $S{=}1$. The x-axis shows the chosen top-$k$ and the y-axis shows $\Delta_m$.
  (b) Bar chart of $\Delta_m$ under fine-grained allocations.  }
  \label{fig:mod_vs_ase_all}
  \vspace{-4mm}
\end{figure}

\textbf{Fine-grained expert settings.} We further adopt the fine-grained settings following Sec.~\ref{sec:exp_setting} for both baseline and ASE. As illustrated in Fig. 3(b), ASE consistently improves over the baseline across all settings, with the most pronounced gain observed in the 64-expert configuration. In addition, finer expert granularity yields nearly linear improvements, highlighting the effectiveness of fine-grained designs in facilitating task cooperation and expert specialization, particularly when combined with ASE.

\textbf{Results on PASCAL-Context.} 
Table~\ref{tab:pascal_context_results} summarizes the results of different MTL methods. 
We compare against classical ResNet-18 based methods, including MTAN~\cite{mtan}, NDDR-CNN~\cite{nddr_cnn}, and Cross-Stitch~\cite{cross_sritch}, as well as ViT-based multi-task baselines and our proposed LoRA-MoE variants. 
Importantly, all methods are trained under the same setting with an input resolution of $224\times224$ and 40 training epochs to ensure fairness. 

Compared with the vanilla MTL-ViT, the LoRA-MoE baselines show clear improvements across tasks, e.g., \emph{Seg.} increases from 69.1 to 73.8 mIoU and $\Delta_m$ rises from $+2.68\%$ to over $+6\%$. Building on this, our proposed ASE further improves performance. 
Specifically, under $(32/6/2/2)$ configuration, ASE achieves the best overall results with 74.0 mIoU on \emph{Seg.}, 60.3 mIoU on \emph{H. Parts}, and an average $\Delta_m$ of \textbf{+7.58\%}, demonstrating consistent gains over the baseline. Notably, these improvements are achieved with only a marginal $\sim$4\% increase in parameters over the plain ViT baseline, highlighting the efficiency of the LoRA-MoE design in delivering substantial performance gains without significant model size overhead.

When compared with ResNet-18 based methods, our approach shows a clear advantage. 
For example, Cross-Stitch~\cite{cross_sritch} improves ResNet-18 to $\Delta_m=+0.66\%$, whereas our ASE achieves more than $+7\%$ under the same unified setting. 

\begin{figure}[t]
\vspace{-3mm}  
      \centering
      \includegraphics[width=\linewidth]{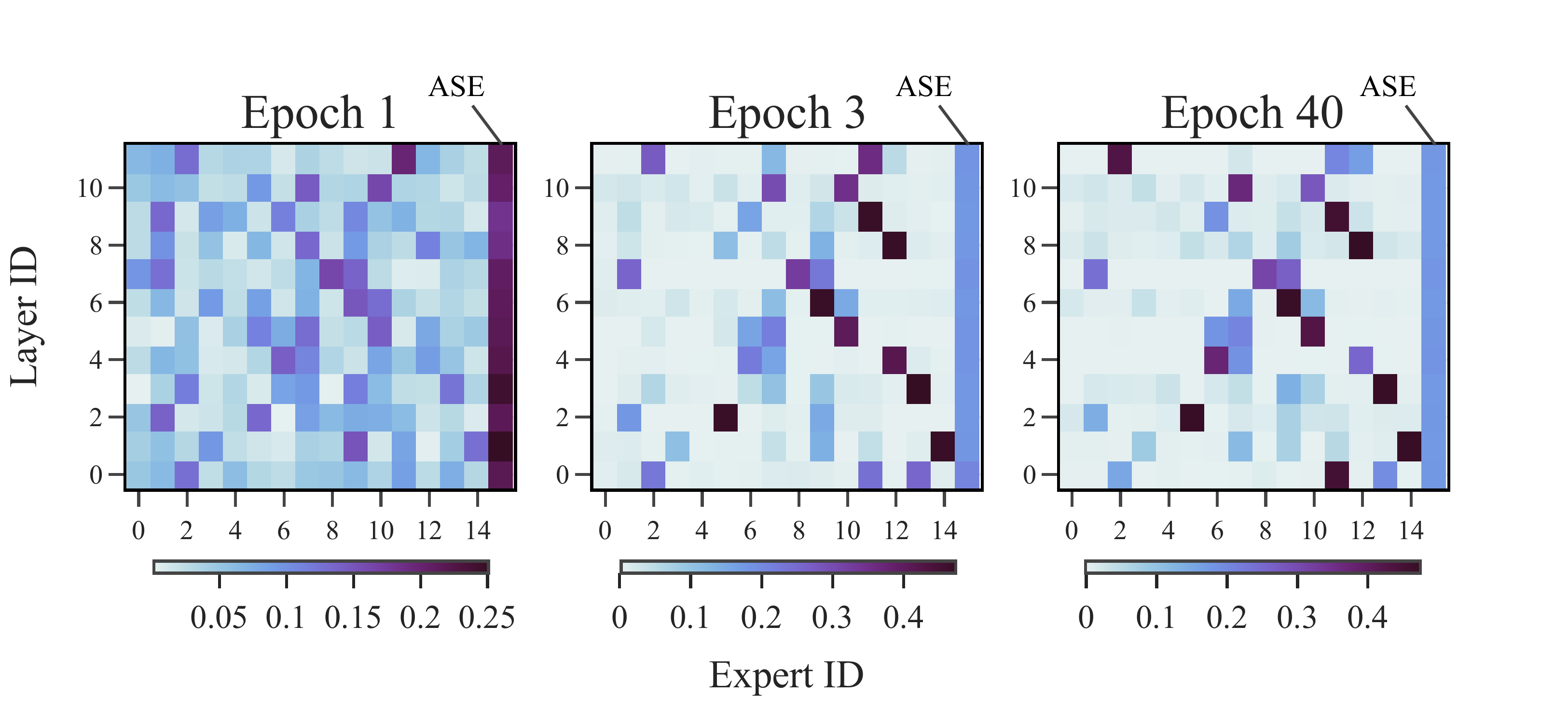}
       \vspace{-8mm}
      \caption{\textbf{Activation frequency of experts across layers.} We visualize the activation frequency of experts for task \emph{Norm.} across all layers at epochs 1, 3, and 40 (out of 40). Here, the y-axis represents the layer index, and the x-axis represents the 16 experts, including one ASE. The color bar indicates the normalized frequency. }
      \label{fig:placeholder}
      \vspace{-3mm}
  \end{figure}
  
\textbf{Visualization of the Activation Frequency of Experts.} 
Fig.~\ref{fig:placeholder} illustrates the activation frequency of experts across layers for task \emph{Norm.}.
Specifically, at the early training stage (epoch 1), ASE dominates most layers, emphasizing its role in bridging the STL-to-MTL transition.
As training progresses (epoch 3), the distribution of sparse experts becomes sharper and increasingly task-specific, while ASE gradually reduces its contribution to alleviate gradient conflicts. 
By the final stage (epoch 40), expert selection stabilizes, exhibiting clear specialization. This dynamic transition demonstrates that ASE adaptively adjusts its influence throughout different training phases, thereby facilitating effective early knowledge transfer and mitigating later gradient conflicts. 
Moreover, the joint normalization of ASE ensures a stable balance between sparse and shared experts, as well as the output of the LoRA-experts and the frozen weights.

\section{Conclusion}
In this paper, we introduced adaptive shared experts (ASE)  within a LoRA-MoE framework to address the limitations of naive shared experts in MoE-based MTL, facilitating both STL-to-MTL transition and effective expert specialization and cooperation. In addition, we incorporated the notion of fine-grained experts into MoE for MTL. Extensive experiments on PASCAL-Context demonstrated the robustness and effectiveness of ASE across different settings, as well as the benefits of fine-grained configurations for improving multi-task performance.
\bibliographystyle{IEEEbib}
\bibliography{reference}

\end{document}